\documentclass[10pt,twocolumn,letterpaper]{article}
\usepackage{cvpr}
\usepackage{times}
\usepackage{epsfig}
\usepackage{graphicx}
\usepackage{amsmath}
\usepackage{amssymb}
\usepackage{verbatim}
\usepackage{mathtools}
\usepackage{dcolumn}
\usepackage[utf8]{inputenc}
\cvprfinalcopy % *** Uncomment this line for the final submission

\ifcvprfinal\fi
\begin{document}

%%%%%%%%% TITLE
\title{Spatio-Temporal Dynamics and Semantic Attribute Enriched\\ Visual Encoding for Video Captioning}
\vspace{-3mm}
%%%%%%%%% AUTHORS
\author{Nayyer Aafaq \hspace{4mm} Naveed Akhtar  \hspace{4mm} Wei Liu \hspace{4mm} Syed Zulqarnain Gilani \hspace{4mm} Ajmal Mian\\
Computer Science and Software Engineering,\\
The University of Western Australia.\\
{\tt\small nayyer.aafaq@research.uwa.edu.au,} \hspace{-2mm}
{\tt\small \{naveed.akhtar,\hspace{-1mm} wei.liu,\hspace{-1mm} syed.gilani,\hspace{-1mm} ajmal.mian\}@uwa.edu.au}
}

\maketitle
%\thispagestyle{empty}

%%%%%%%%% ABSTRACT
\begin{abstract}
  Automatic generation of video captions is a fundamental challenge in computer vision. Recent techniques typically employ a combination of Convolutional Neural Networks (CNNs) and Recursive Neural Networks (RNNs) for video captioning. These methods mainly focus on tailoring sequence learning through RNNs for better caption generation, whereas off-the-shelf visual features are borrowed from CNNs. We argue that careful designing of visual features for this task is equally important, and present a visual feature encoding technique to generate semantically rich captions using Gated Recurrent Units (GRUs). Our method embeds rich temporal dynamics in visual features by hierarchically applying  Short Fourier Transform to CNN features of the whole video. It additionally derives high level semantics from an object detector to enrich the representation with spatial dynamics of the detected objects. The final representation is projected to a compact space and fed to a language model. By learning a relatively simple language model comprising two GRU layers, we establish new state-of-the-art on MSVD  and MSR-VTT datasets for METEOR and ROUGE$_L$ metrics.  
\end{abstract}

%%%%%%%%% BODY TEXT
\vspace{-3mm}
\section{Introduction}
\label{sec:Intro}
Describing videos in natural language is  trivial for humans, however it is a very complex task for machines. To generate meaningful video captions, machines are required to understand objects, their interaction, spatio-temporal order of events and other such minutiae in videos; yet, have the ability to articulate these details in grammatically correct and meaningful natural language sentences~\cite{aafaq2018videosurvey}. 
The bicephalic nature of this problem has recently led researchers from Computer Vision and Natural Language Processing (NLP) to combine efforts in addressing its challenges~\cite{languageandvisioncvpr2015, languageandvisioncvpr2018, languageandvisioniccv2015, languageandvisionnaacl2018}.
Incidentally, wide applications of video captioning in emerging technologies, e.g.~procedure generation from instructional videos~\cite{alayrac2016unsupervised}, video indexing and retrieval~\cite{song2018quantization, wang2018survey}; have recently caused it to  receive attention as a fundamental task in Computer Vision. 

Early methods in video captioning and description, e.g.~\cite{kojima2002natural,barbu2012video} primarily aimed at generating the correct Subject, Verb and Object (a.k.a.~SVO-Triplet) in the captions. More recent methods~\cite{venugopalan2015sequence,rohrbach2017movie} rely on Deep Learning \cite{lecun2015deep} to build frameworks resembling a typical neural machine translation system that can generate a single sentence~\cite{xu2016msr,pan2016jointly} or multiple sentences~\cite{rohrbach2014coherent,  shin2016beyond, yu2016video}  to describe  videos. 
The two-pronged problem of video captioning provides a default division for the deep learning  methods to encode visual contents of videos using Convolutional Neural Networks (CNNs)~\cite{simonyan2014very, tran2015learning} and decode those into captions using language models. Recurrent Neural Networks (RNNs)~\cite{elman1990finding, cho2014learning, hochreiter1997long} are the natural choice for the latter component of the problem.   

Since semantically correct sentence generation has a longer history in the field of NLP, deep learning based captioning techniques mainly focus on language modelling~\cite{venugopalan2014translating, Pan_2017_CVPR}.  %learning semantic attributes  either from the captions themselves or from off-the-shelf visual features~\cite{Pan_2017_CVPR, gan2017semantic}.
For visual encoding, these methods forward pass video frames through a pre-trained 2D CNN; or a video clip through a 3D CNN, and extract features from an inner layer of the network - referred as  `extraction layer'. 
Features of frames/clips are commonly combined with mean pooling to compute the final representation of the whole video.
This, and similar other visual encoding techniques~\cite{pan2016jointly, venugopalan2014translating, gan2017semantic, Pan_2017_CVPR}  - due to the nascency of video captioning research - grossly under-exploit the prowess of visual representation for the captioning task. To the best of our knowledge, this paper presents the first work that concentrates on improving the visual encoding mechanism for the captioning task.     

\begin{figure*}[t] %  figure placement: here, top, bottom, or page
  \centering
     \includegraphics[width=5.7in]{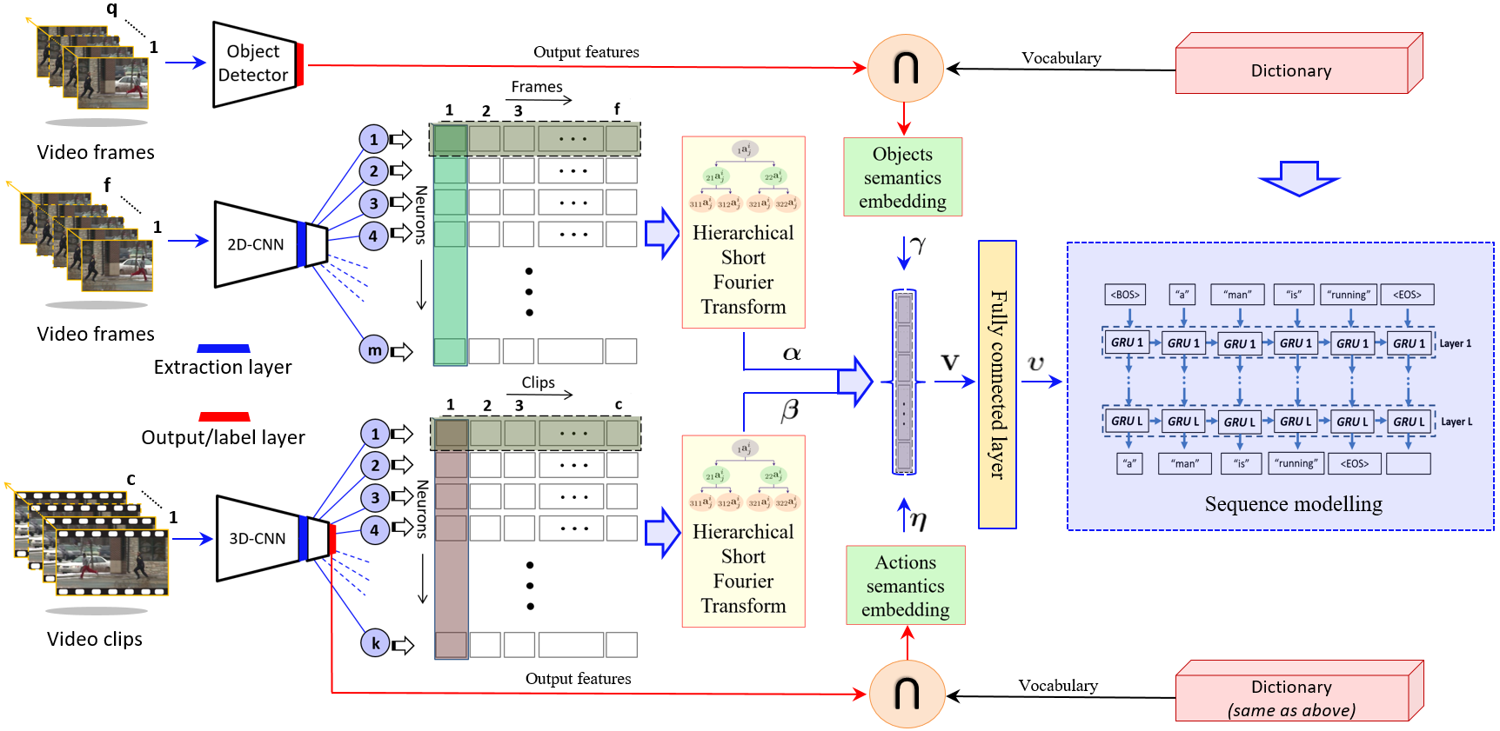}
     \caption{The `$c$' clips and `$f$' frames of a video are processed with 3D and 2D CNNs respectively. Neuron-wise Short Fourier Transform is applied hierarchically to the extraction layer activations of these networks (using the whole video). This results in spatio-temporal dynamics enriched encodings  $\boldsymbol{\alpha}$ and $\boldsymbol{\beta}$. Relevant high-level object semantics $\boldsymbol\gamma$ and action semantics $\boldsymbol\eta$ are derived using the intersection of vocabulary from the language model dictionary with the labels of 3D CNN and an Object Detector. The output features of the Object Detector are also used to embed spatial dynamics of the scene and plurality of the objects therein. The resulting codes are compressed with a fully-connected layer and used to learn a multi-layer GRU as a language model.}
  \label{fig:framework}
  \vspace{-3mm}
\end{figure*}

We propose a visual encoding technique to compute representations enriched with spatio-temporal dynamics of the scene, while also accounting for the high-level semantic attributes of the videos.
Our visual code (`${\bf v}$' in Fig.~\ref{fig:framework}) fuses information from multiple sources. We process activations of 2D and 3D CNN extraction layers by hierarchically applying Short Fourier Transform~\cite{oppenheim1999discrete} to them, where  InceptionResNetv2~\cite{szegedy2017inception} and C3D~\cite{tran2015learning} are used as the 2D and 3D CNNs respectively.  The proposed neuron-wise activation transformation using \textit{whole videos} results in encoding fine temporal dynamics of the scenes.   
We encode spatial dynamics by processing objects' locations and their multiplicity information extracted from an Object Detector (YOLO~\cite{redmon2017yolo9000}). The semantics attached to the output layers of the Object Detector and the 3D CNN are also exploited to embed high-level semantic attributes in our visual codes.
We compress the visual codes and learn a language model using the resulting representation. With highly rich visual codes, a relatively simple Gated Recurrent Unit (GRU) network is proposed for language modeling, comprising two layers, that already results in on-par or better performance compared to the existing sophisticated models~\cite{wang2018reconstruction, wang2018m3, Pan_2017_CVPR, gan2017semantic} on multiple evaluation metrics.  

The main contributions of this paper are as follows. We propose a visual encoding technique that effectively encapsulates spatio-temporal dynamics of the videos and embeds relevant high-level semantic attributes in the visual codes for video captioning. The proposed visual features contain the detected object attributes, their frequency of occurrences as well as the evolution of their locations over time. We establish the effectiveness of the proposed encoding by learning a GRU-based language model and perform thorough experimentation on  MSVD~\cite{chen2011collecting} and MSR-VTT~\cite{xu2016msr} datasets. Our method  achieves up to $2.64\%$ and $2.44\%$ gain in the state-of-the-art on METEOR and ROUGE$_L$ metrics for these datasets.  % dataset, and also outperforms the nearest competitor on MSR-VTT for the same metric. %, while maintaining highly competitive results on CIDER$_D$ and BLEU-4 . 

%-------------------------------------------------------------------------
\vspace{-2mm}
\section{Related Work}
Classical methods in video captioning commonly use template based techniques in  which Subject (S), Verb (V), and Object (O)  are detected separately and then joined together in a sentence.  
%Video captioning was, and to some extent is still, done using the classical template based approach in which Subject (S), Verb(V), Object (O) and Place (P) are detected separately and then joined together in a sentence template. 
However, the advancement  of deep learning research has also transcended to modern video captioning methods. The latest approaches in this direction generally exploit deep learning for visual feature encoding as well as its decoding into meaningful captions. %using 2D/3D-CNN and LSTM/GRU to learn the sequence. 

In template based approaches, the first successful video captioning method was proposed by Kojima et al.~\cite{kojima2002natural} that focuses  on describing videos of one person performing one action only. Their heavy reliance on the correctness of manually created activity concept hierarchy and state transition model prevented its extension to more complex videos. Hanckmann et al.~\cite{hanckmann2012automated} proposed a method to automatically describe events involving multiple actions (seven on average), performed by one or more individuals. Whilst most of the prior work was restricted to constrained domains~\cite{khan2011human, barbu2012video}, Krishnamoorthy et al.~\cite{krishnamoorthy2013generating} led the early works of describing open domain videos. \cite{guadarrama2013youtube2text} proposed semantic hierarchies to establish relationships between actor, action and objects. \cite{rohrbach2013translating} used CRF to model the relationship between visual entities and treated video description as a machine translation problem. However, the aforementioned approaches depend on predefined sentence templates and fill in the template by detecting entities from classical methods. These approaches are not sufficient for the syntactically rich sentence generation to describe open domain videos. 

In contrast to the methods mentioned above, deep models directly generate sentences given a visual input. For example LSTM-YT~\cite{venugopalan2014translating} feed in visual contents of video obtained by average pooling all the frames into LSTM and produce the sentences. LSTM-E~\cite{pan2016jointly} explores the relevance between the visual context and sentence semantics. The initial visual features in this framework were obtained using 2D-CNN and 3D-CNN whereas the final video representation was achieved by average pooling the features from frames / clips neglecting the temporal dynamics of the video. TA~\cite{yao2015describing} explored the temporal domain of video by introducing an attention mechanism to assign weights to the features of each frame and later fused them based on attention weights. S2VT~\cite{venugopalan2015sequence} incorporated optical flow to cater for the temporal information of the video. SCN-LSTM~\cite{gan2017semantic} proposed semantic compositional network that can detect the semantic concepts from mean pooled visual content of the video and fed that information into a language model to generate captions with more relevant words. LSTM-TSA\cite{Pan_2017_CVPR} proposed a transfer unit that extracts semantic attributes from both images as well as mean pooled visual content of videos and added it as a complementary information to the video representation to further improve the quality of caption generation. M$^3$-VC~\cite{wang2018m3} proposed a multi-model memory network to cater for long term visual-textual dependency and to guide the visual attention.

Even though the above methods have employed deep learning, they have used mean pooled visual features or attention based high level features from CNNs. These features have been used directly in their framework in the language model or by introducing additional unit in the standard framework. We argue that this technique under-utilizes the state of the art CNN features in video captioning framework. We propose features that are rich in visual content and empirically show that this enrichment of visual features alone when combined with a standard and simple language model can outperform existing state of the art methods. Visual features are part of every video captioning framework. Hence, instead of using high level or mean pooled features, building on top of our visual features can further enhance the video captioning frameworks' performances. 
% --------------------------------------------------------------------------------------
\section{Proposed Approach}
Let $\mathcal V$ denote a video that has `$f$' frames or `$c$' clips. The fundamental task in automatic video captioning is to generate a  textual sentence $\mathcal S = \{\mathcal W_1, \mathcal W_2,..., \mathcal W_w\}$ comprising `$w$' words that matches closely to human generated captions for the same video. Deep learning based video captioning methods typically define an energy loss function of the following form for this task:
\begin{align}
    \Xi ({\bf v}, \mathcal S) = - \sum \limits_{t = 1}^w  \log \text{Pr} \left( \mathcal W_t | {\bf v}, \mathcal W_1,...\mathcal W_{t-1} \right),
    \label{eq:prob}
\end{align}

% \begin{align}
%     \Xi ({\bf v}, \mathcal S) = - \sum \limits_{t = 1}^w  \log \text{Pr} \left( \mathcal W_t | {\bf v}, \mathcal W_1,...\mathcal W_{t-1} \right),
%     \label{eq:prob}
% \end{align}
where Pr(.) denotes the probability, and ${\bf v} \in \mathbb R^{d}$ is a visual representation of $\mathcal V$.
By minimizing the cost defined as the  Expected value of the energy $\Xi(.)$ over a large corpus of videos, it is hoped that the inferred model $\mathcal M$ can automatically generate meaningful captions for  unseen videos.

In this formulation, `${\bf v}$' is considered a training input, that makes remainder of the problem a sequence learning task.
%In Eq.~(\ref{eq:prob}) the visual feature ${\bf v}$ is expected as an input, whereas the underlying objective function is best suited for sequence learning.
Consequently, the existing methods in video captioning mainly focus on tailoring RNNs~\cite{elman1990finding} or LSTMs~\cite{hochreiter1997long} to generate better captions, \textit{assuming} effective visual encoding of $\mathcal V$ to be available in the form of `${\bf v}$'. The representation prowess of CNNs has made them the default choice for visual encoding in the existing literature. However, due to the nascency of video captioning research, only primitive methods of using CNN features for  `${\bf v}$' can be found in the literature.
These methods directly use  2D/3D CNN features or their concatenations for visual encoding, where the temporal dimension of the video is resolved by mean pooling~\cite{pan2016jointly, Pan_2017_CVPR, gan2017semantic}. 
%Thus, the existing methods focus on $\mathcal M(.)$ in the mapping $\mathcal M({\bf v}) \rightarrow \mathcal S$.

We acknowledge the role of apt sequence modeling for video description, however, we also argue that designing specialized visual encoding techniques for captioning is equally important. Hence, % in this first of its kind contribution\footnote{To the best of our knowledge.}, 
we mainly focus on the operator $\mathcal Q(.)$ in the mapping $\mathcal M(\mathcal Q( \mathcal V))) \rightarrow \mathcal S$, where $\mathcal Q( \mathcal V) \rightarrow {\bf v}$. We propose a visual encoding technique that, along harnessing the power of CNN features, explicitly encodes spatio-temporal dynamics of the scene in the visual representation, and embeds semantic attributes in it to further help the sequence modelling phase of video description to generate semantically rich textual sentences.

\subsection{Visual Encoding}
\label{sec:VE}
For clarity, we describe the visual representation of a video $\mathcal V$ as ${\bf v} = [\boldsymbol{\alpha};~\boldsymbol{\beta}; ~\boldsymbol{\gamma};~\boldsymbol{\eta}]$, where $\boldsymbol{\alpha}$ to $\boldsymbol{\gamma}$ are themselves column-vectors computed by the proposed technique.
We explain these computations in the following. 

\vspace{-2mm}
\subsubsection{Encoding Temporal Dynamics}
\label{sec:TE}
In the context of video description, features extracted from pre-trained 2D-CNNs, e.g.~VGG~\cite{simonyan2014very} and 3D-CNNs, e.g.~C3D~\cite{tran2015learning} have been shown useful for visual encoding of videos. The standard practice is to forward pass individual video frames through a 2D CNN and store  activation values of a pre-selected \textit{extraction} \textit{layer} of the network.
Then, perform mean pooling over those  activations for all the frames to compute the visual representation.
A similar procedure is adopted with 3D CNN with a difference that video clips are used in forward passes instead of   frames. %A simplistic concatenation of the resulting 2D-3D features is also common in the literature.

%It is also a known fact that concatenating  features from 2D-3D CNNs further improves captioning performance. 

\newcommand\myATwoD{\stackrel{\mathclap{\normalfont\tiny\mbox{2D}}}{a}}
\newcommand\myAThreeD{\stackrel{\mathclap{\normalfont\tiny\mbox{3D}}}{a}}

A simple mean pooling operation over activation values is bound to fail in encoding  fine-grained temporal dynamics of the  video.
This is true for both 2D and 3D CNNs, despite the fact that the latter models video clips.
We address this shortcoming by defining  transformations $\mathcal T_f (\mathcal F) \rightarrow \boldsymbol{\alpha}$ and $\mathcal T_c (\mathcal C) \rightarrow \boldsymbol{\beta}$, such that $\mathcal F = \{ {\bf \myATwoD}_1, {\bf \myATwoD}_2,..., {\bf \myATwoD}_f\}$ and $\mathcal C = \{ {\bf \myAThreeD}_1, {\bf \myAThreeD}_2,..., {\bf \myAThreeD}_c\}$. Here, ${\bf \myATwoD}_t$ and ${\bf \myAThreeD}_t$ denote the activation vectors of the extraction layers of 2D and 3D CNNs for the $t^{\text{th}}$ video frame and video clip respectively. The aim of these transformations is to compute $\boldsymbol{\alpha}$ and $\boldsymbol{\beta}$ that encode temporal dynamics of the \textit{complete} video with high fidelity. 

We use the last $avg~pool$ layer of InceptionResnetV2~\cite{szegedy2017inception} to compute ${\bf \myATwoD}_i$, and the $fc6$ layer of C3D~\cite{tran2015learning} to get ${\bf \myAThreeD}_i$. The transformations $\mathcal T_{f/c}(.)$ are defined over the activations of those extraction layers. Below, we explain $\mathcal T_{f}(.)$ in detail. The transformation  $\mathcal T_{c}(.)$ is similar, except that it uses activations of clips instead of   frames. 

\begin{figure}[t] %  figure placement: here, top, bottom, or page
  \centering
     \includegraphics[width=0.4\textwidth]{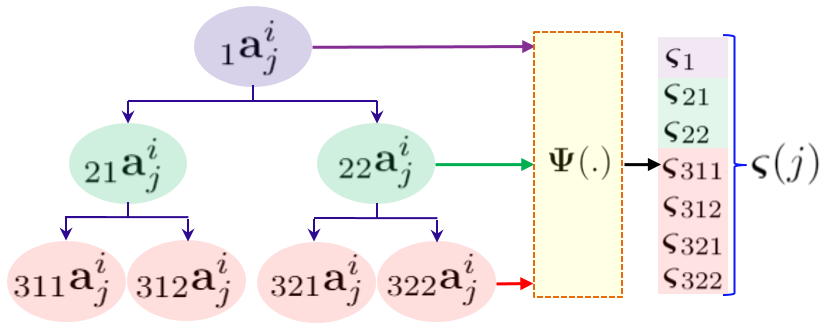}
     \caption{Illustration of hierarchical  application of Short Fourier Transform $\boldsymbol\Phi(.)$ to the activations ${\bf a}_j^i$  of the $j^{\text{th}}$ neuron of the extraction layer for the $i^{\text{th}}$ video.  }
  \label{fig:FTP}
  \vspace{-6mm}
\end{figure}

Let $a_{j,t}^i$ denote the activation value of the $j^{\text{th}}$ neuron of the network's extraction layer for the $t^{\text{th}}$ frame of the $i^{\text{th}}$ training video. We leave out the superscript 2D for better readability. To perform the transform, we first define ${}_1{\bf a}_j^i = [a_{j,1}^i, a_{j,2}^i,...,a_{j,f}^i] \in \mathbb R^f$ and compute $\Psi({}_1{\bf a}_j^i) \rightarrow \boldsymbol{\varsigma}_1 \in \mathbb R^ p$, where the operator $\boldsymbol{\Psi}(.)$ computes the Short Fourier Transform~\cite{oppenheim1999discrete} of the vector in its argument and stores the first `$p$' coefficients of the transform. Then, we divide ${}_1{\bf a}_j^i$ into two smaller vectors ${}_{21}{\bf a}_j^i \in \mathbb R^{h}$ and ${}_{22}{\bf a}_j^i \in \mathbb R^{h-f}$, where $h = \lfloor \frac{f}{2} \rfloor$. We again apply the operator  $\boldsymbol{\Psi}(.)$ to these vectors to compute $\boldsymbol\varsigma_{21}$ and $\boldsymbol\varsigma_{22}$ in p-dimensional space. We recursively perform  the same operations on $\boldsymbol\varsigma_{21}$ and $\boldsymbol\varsigma_{22}$ to get the p-dimensional vectors $\boldsymbol\varsigma_{311}$, $\boldsymbol\varsigma_{312}$ , $\boldsymbol\varsigma_{321}$, and $\boldsymbol\varsigma_{322}$. We combine all these vectors as $\boldsymbol\varsigma (j) = [\boldsymbol\varsigma_{1}, \boldsymbol\varsigma_{21}, \boldsymbol\varsigma_{22}, ..., \boldsymbol\varsigma_{322}] \in \mathbb R^{(p \times 7) \times 1}$. We also illustrate this operation in Fig.~\ref{fig:FTP}. The same operation is performed individually for each neuron of our extraction layer. We then concatenate $\boldsymbol\varsigma (j): j \in \{1,2,...,m\}$ to form $\boldsymbol{\alpha} \in \mathbb R^{(p \times 7 \times m) \times 1}$, where $m$ denotes the number of neurons in the  extraction layer.  As a result of performing   $\mathcal T_f(\mathcal F) \rightarrow \boldsymbol{\alpha}$, we have computed a representation the video while accounting for fine temporal dynamics in the whole sequence of video frames.     
Consequently,  $\mathcal T_f(.)$ results in a much more informative representation than that obtained with mean pooling of the neuron activations.

We define $\mathcal T_{c}(.)$ in a similar manner for the set $\mathcal C$ of video clip activations. This transformation results in $\boldsymbol{\beta}\in \mathbb R^{(p \times 7 \times k) \times 1}$, where $k$ denotes the number of neurons in the extraction layer of the 3D CNN. It is worth mentioning that a 3D CNN is already trained on short video \textit{clips}. Hence, its features  account for the temporal dimension of $\mathcal V$ to some extent. Nevertheless, accounting for the fine temporal details in the whole video  adds to our  encoding significantly (see Section~\ref{sec:Exp}). It is noteworthy that exploiting  Fourier Transform in a hierarchical fashion to encode temporal dynamics has also been considered in human action recognition~\cite{wang2014learning, rahmani20163d}. However, this work is the first to apply Short Fourier Transform hierarchically for video captioning.   

\vspace{-2mm}
\subsubsection{Encoding Semantics and Spatial Evolution}
\label{sec:SSE}
It is well-established that the latter layers of CNNs are able to learn features at higher levels of abstraction due to hierarchical application of convolution operations in the earlier layers~\cite{lecun2015deep}. The common use of activations of e.g.~fully-connected layers as visual features for captioning is also motivated by the fact that these representations are  \textit{discriminative transformations of high-level} video features. We take this concept further and argue that the output layers of CNNs can themselves serve as discriminative encodings of the highest abstraction level for video captioning. % - the level that we encounter in video captions. 
We describe the  technique to effectively exploit these features  in the paragraphs to follow. Here, we briefly emphasize that the output layer of a network contains additional  information for video captioning beyond what is provided by the commonly used extraction layers of networks, because:
\begin{enumerate}
\item The output labels are yet another transformation of the extraction layer features, resulting from network weights that are  unaccounted for by  extraction layer.
\item  The semantics attached to the output layer are at the same level of abstraction that is encountered in video  captions - a unique property of the output layers. 
\end{enumerate}

We  use  the output layers of an Object Detector (i.e.~YOLO~\cite{redmon2017yolo9000}) and a 3D CNN (i.e.~C3D~\cite{tran2015learning}) to extract semantics pertaining to the objects and actions recorded in videos. The core idea is to quantitatively embed object labels, their frequencies of occurrence, and evolution  of their spatial locations in videos   in the visual encoding vector. Moreover, we also aim to enrich our visual encoding with the semantics of actions performed in  the video. The details of materializing  this concept are presented below. 
\vspace{-2mm}
\paragraph{Objects Information:}
Different from classifiers that only predict labels of input images/frames, object detectors can localize multiple objects in individual frames, thereby providing cues for ascertaining  plurality of the same type of objects in individual frames and  evolution of objects' locations in multiple frames. Effective embedding of such high-level information in vector `${\bf v}$' promises descriptions that can clearly differentiate between e.g.~`people running' and `person walking' in a video. 

The sequence modeling component of a video captioning system generates a textual sentence  by selecting words from a large dictionary $\mathcal D$. An object detector provides a set $\widetilde{\mathcal L}$ of object labels at its output. We first compute $\mathcal L =  \mathcal D \bigcap\widetilde{\mathcal L}$, and define $\boldsymbol{\gamma} = [\boldsymbol\zeta_1, \boldsymbol\zeta_2, ..., \boldsymbol\zeta_{|\mathcal L|} ]$, where $|.|$ denotes the cardinality of a set.
The vectors $\boldsymbol{\zeta}_i, \forall i$ in $\boldsymbol{\gamma}$ are further defined with the help `$q$' frames sampled from the original video. We perform this sampling using a fixed time interval between the sampled frames of  a given video. The samples are passed through the object detector and its output is utilized in computing $\boldsymbol{\zeta}_i, \forall i$. %Note that, besides simply providing object labels, an output layer of an object detector also provides us with the location of the object appearing in a frame. This allows us 
A vector $\boldsymbol{\zeta}_i$ is  defined as $\boldsymbol{\zeta}_i = [\text{Pr}(\ell_i), \text{Fr}(\ell_i), {\boldsymbol{\nu}}_i^1,  {\boldsymbol{\nu}}_i^2,...,  {\boldsymbol{\nu}}_i^{(q-1)}]$, where $\ell_i$ indicates the $i^{\text{th}}$ element of $\mathcal L$ (i.e.~an object name), Pr(.) and Fr(.) respectively compute the probability and frequency of occurrence of the object corresponding to $\ell_i$, and $\boldsymbol{\nu_i}^z$ represent the velocity of the object between the frames $z$ and $z+1$ (in the sampled $q$ frames).

We define $\boldsymbol{\gamma}$  over `$q$' frames, whereas the used object detector processes individual frames that results in a probability and frequency value for each frame. We resolve this and related  mismatches by using the following definitions of the components of $\boldsymbol{\zeta}_i$: 
\begin{itemize}
\item Pr(.) $= \max\limits_z  \text{Pr}_z(.)$ : $z \in \{1,...,q\}$.
\item Fr(.) = $\frac{\max\limits_z  \text{Fr}_z(.)}{N}$ : $z \in \{1,...,q\}$, where `$N$' is the allowed  maximum number of the same class of objects detected in a frame. We let $N =10$ in experiments.
\item $\nu_i^z = [\delta_x^z, \delta_y^z]$ : $\delta_x^z = \tilde x^{z+1} - \tilde x^{z}$ and $\delta_y^z = \tilde y^{z+1} - \tilde y^{z}$. Here, $\tilde x, \tilde y$ denote the Expected values of the $x$ and $y$ coordinates of the same type of objects in a given frame, such that the coordinates are also normalized by the respective frame dimensions. 
\end{itemize}

We let $q = 5$ in our experiments, resulting in $\boldsymbol{\zeta}_i \in \mathbb R^{10}, \forall i$ that compose $\boldsymbol{\gamma} \in \mathbb R^{(10 \times |\mathcal L|) \times 1}$.  The indices of coefficients in $\boldsymbol\gamma$ identify the object labels in videos (i.e.~probable nouns to appear in the description). Unless an object is detected in the video, the coefficients of $\boldsymbol{\gamma}$ corresponding to it are kept zero. 
The proposed embedding of  high level semantics in $\boldsymbol{\gamma}$ contain highly relevant information about objects in explicit form for a sequence learning module of video description system. 

\vspace{-2mm}
\paragraph{Actions Information:} Videos generally record  objects and their interaction. The latter is best described by the actions performed in the videos. We already use a 3D CNN that learns action descriptors for the videos. We tap into the output layer of that network to further embed high level action information in our visual encoding. To that end, we compute $\mathcal A = \widetilde{\mathcal A} \bigcap \mathcal D$, where $\mathcal A$ is the set of labels at the output of the 3D CNN. Then, we define $\boldsymbol{\eta} = \left[ [\vartheta_1, \text{Pr}(\ell_1)],   [\vartheta_2, \text{Pr}(\ell_2)], ...,  [\vartheta_{|\mathcal A|}, \text{Pr}(\ell_{|\mathcal A|})]\right] \in \mathbb R^{(2 \times |\mathcal A|) \times 1}$, where $\ell_i$ is the $i^{\text{th}}$ element of $\mathcal A$ (an action label) and $\vartheta$ is a binary variable that is 1 only if the action is predicted by the network. 

We concatenate the above described vectors $\boldsymbol{\alpha}, \boldsymbol{\beta}, \boldsymbol{\gamma}$ and $\boldsymbol{\eta}$ to form our visual encoding vector ${\bf v} \in \mathbb R^d$, where $d = 2\times (p \times 7 \times m) + (10 \times |\mathcal L|) + (2 \times |\mathcal A|)$. Before passing this vector to a sequence modelling component of our method, we perform its compression using a fully connected layer, as shown in Fig.~\ref{fig:framework}. Using $tanh$ activation function and fixed weights, this  layer projects `${\bf v}$' to a 2K-dimensional space. The resulting projection `$\boldsymbol{\upsilon}$'  is used by our language model.  
% \vspace{1mm}
\subsection{Sequence Modelling}
\vspace{1mm}
We follow the common pipeline of video description techniques that feeds visual representation of a video to a sequence modelling component, see Fig.~\ref{fig:framework}. Instead of resorting to a sophisticated language model, we develop a relatively simpler model employing multiple layers of  Gated Recurrent Units (GRUs)~\cite{cho2014learning}.  
GRUs are known to be more robust to vanishing gradient problem - an issue encountered in long captions - due to their ability of remembering the relevant information and forgetting the rest over time.

A GRU has two gates: reset $\Gamma_r$ and update $\Gamma_u$, where the  update gate decides how much the unit updates its previous memory and the reset gate determines how to combine the new input with the previous memory.
Concretely, our language model computes the hidden state $h^{<t>}$ of a GRU as:
%Let $\odot$ denote the element-wise multiplication of two vectors. GRU computes its current hidden state   
\vspace{0.2mm}
$$\Gamma_u = \sigma(W_u[h^{<t-1>}, x^{<t>}] + b_u)$$
\vspace{0.2mm}
$$\Gamma_r = \sigma(W_r[h^{<t-1>}, x^{<t>}] + b_r)$$
\vspace{0.2mm}
$$\tilde h^{<t>} = \tanh{(W_h[\Gamma_r \odot h^{<t-1>}, x^{<t>}] + b_h}$$
\vspace{0.2mm}
$$ h^{<t>} = \Gamma_u \odot \tilde h^{<t>} + (1 - \Gamma_u) \odot h^{<t-1>}$$ 
\vspace{0.2mm}
where, $\odot$ denotes the hadamard product, $\sigma(.)$ is sigmoid activation , $W_q, \forall q$ are learnable weight matrices, and $b_{u/r/h}$ denote the respective biases. In our approach, $h^{<0>} = \boldsymbol{\upsilon}$ for a given video, whereas the signal $x$ is the word embedding vector. 
In Section~\ref{sec:Exp}, we report results using two layers of GRUs, and demonstrate that our language model under the proposed straightforward sequence modelling  already provides highly competitive performance due to the proposed visual encoding. 
% ---------------------------------------------------------------------------------------------

\section{Experimental Evaluation}
\label{sec:Exp}

\vspace{1mm}
\subsection{Datasets}
We evaluate our technique using two popular benchmark datasets from the existing literature in video description, namely Microsoft Video Description (MSVD) dataset~\cite{chen2011collecting}, and MSR-Video To Text (MSR-VTT) dataset~\cite{xu2016msr}.  We first give details of these datasets and their processing performed in this work, before discussing the experimental results.  

\vspace{4mm}
\noindent \textbf{MSVD Dataset}~\cite{chen2011collecting}: This dataset is composed of  1,970 YouTube open domain videos that predominantly show only a single activity each. Generally, each clip is spanning  over 10 to 25 seconds. The dataset provides multilingual human annotated sentences as captions for the videos. We experiment with the captions in English. On average, 41 ground truth captions can be associated with a single video. For benchmarking, we follow the common data split of 1,200 training samples, 100 samples for validation and 670 videos for testing~\cite{yao2015describing, wang2018m3, gan2017semantic}. 

\vspace{2mm}
\noindent\textbf{MSR-VTT Dataset}~\cite{xu2016msr}:
This recently introduced open domain videos dataset contains a wide variety of videos for the captioning task. It consists of 7,180 videos that are transformed into 10,000 clips. The clips are grouped into 20 different categories. Following the common settings~\cite{xu2016msr}, we divide the 10,000 clips into 6,513 samples for training, 497 samples for validation and the remaining 2,990 clips for testing. Each video is described by 20 single sentence annotations by Amazon Mechanical Turk  (AMT) workers. This is one of the largest clips-sentence pair dataset available for the video captioning task, which is the main  reason of choosing this dataset for benchmarking our technique. 
%---------------------------------------------------------

\vspace{-2mm}
\subsection{Dataset Processing \& Evaluation Metrics} 

%\noindent \textbf{Descriptions Preprocessing}
We converted the captions in both datasets to lower case, and removed all punctuations. All the sentences were then tokenized. We set the vocabulary size for MSVD to  9,450 and for MSR-VTT to 23,500. We employed  ``\textit{fasttext}``~\cite{bojanowski2016enriching} word embedding vectors of dimension 300. Embedding vectors of 1,615 words for MSVD and 2,524 words for MSR-VTT were not present in the pretrained set. Instead of using randomly initialized vectors or ignoring the out of vocabulary words entirely in the training set, we generated embedding vectors for these words using character n-grams within the word, and summing the resulting vectors to produce the final vector. We performed dataset specific fine-tuning on the pretrained word embeddings.

%\noindent \textbf{Evaluation Metrics} 
In order to compare our technique with the existing methods, we report results on the four most popular metrics, including; Bilingual Evaluation Understudy (BLEU)~\cite{papineni2002bleu}, Metric for Evaluation of Translation with Explicit Ordering
(METEOR)~\cite{lavie2005meteor}, Consensus based Image Description Evaluation (CIDEr$_D$)~\cite{vedantam2015cider} and Recall Oriented
Understudy of Gisting Evaluation (ROUGE$_L$)~\cite{lin2004rouge}. We refer to the original works for the concrete definitions of these metrics. The subscript `$D$' in CIDEr indicates the metric variant that inhibits higher values for inappropriate captions in human judgment. Similarly, the subscript `$L$' indicates the variant of ROUGE  that is based on recall-precision scores of the longest common sequence between  the prediction and the ground truth.
%These are among the most commonly used metrics in image and video captioning. 
We used the Microsoft COCO server~\cite{chen2015microsoft} to compute our results. 

%----------------------------------------------------------
% \vspace{-1mm}
\subsection{Experiments}
\label{sec:Exp}
In our experiments reported below\footnote{Due to through evaluation, supplementary material also contains further results. Only the best performing setting  is discussed here.}, we use  InceptionResnetV2 (IRV2)~\cite{szegedy2017inception} as the 2D CNN, whereas C3D~\cite{tran2015learning} is  used as the 3D CNN.
The last `$avg~pool$' layer of the former, and the `$fc6$' layer of the latter are considered as the \textit{extraction layers}.
The 2D CNN is pre-trained on the popular ImageNet dataset~\cite{russakovsky2015imagenet}, whereas Sports 1M dataset~\cite{karpathy2014large} is used for the pre-training of C3D. To process videos, we re-size the frames to match the input dimensions of these networks.
For the 3D CNN, we use 16-frame clips as inputs with an 8-frame overlap. 
YOLO~\cite{redmon2017yolo9000} is used as the object detector in all our experiments. 
To train our language model, we include a start and an end token to the captions to deal with the dynamic length of different sentences.
We set the maximum sentence length to 30 words in the case of experiments with MSVD dataset, and to 50 for the MSR-VTT dataset. 
These length limits are based on the available captions in the datasets. We truncate a sentence if its length exceeds the set limit, and zero pad in the case of shorter length. 

%------------------------------------------------

 We tune the hyper-parameters of our language model on the validation set. The results below use two layers of GRUs, that employ 0.5 as the dropout value. We use the RMSProp algorithm with a learning rate $2 \times 10^{-4}$ to train the models. A batch size of 60 is used for training in our experiments. We performed training of our models for 50 epochs. We used the sparse cross entropy loss to train our model.
 The training is conducted using NVIDIA Titan XP 1080 GPU. We used TensorFlow framework for development.

\vspace{-3mm}
\subsubsection{Results on MSVD dataset}
\vspace{-1mm}
%\subsection{Quantitative results}
%As MSVD is widely reported dataset, we compare our results with top 17 state of the art methods. However, MSR-VTT is relatively new and largest dataset that is not reported much so we take fewer but try to compare with latest approaches reporting on the dataset. 
%\subsubsection{Experimental Results on MSVD}
We comprehensively benchmark our method against the current state-of-the-art in video captioning. We report the results of the existing methods and our approach in Table.~\ref{tab:MSVD}.  For the existing techniques, recent best performing methods are chosen and their results are directly taken from the existing literature (same evaluation protocol is ensured).
The table columns present scores for the metrics BLEU-4 (B-4), METEOR (M), CIDEr$_D$ (C) and ROUGE$_L$ (R). 
%For comprehensive comparison and analysis we compare our results on this dataset with state of the art methods including some popular baseline methods and some latest state of the art published works. The experimental results in terms of BLEU-4 (B-4), METEOR (M), CIDEr (C) and ROUGE$_L$ are shown in Table~\ref{tab:MSVD}. Here we report the best results in terms of METEOR (M) and ROUGE$_L$ as compared to top published state of the art methods and second best results in terms of CIDEr.

The last seven rows of the Table report results of different variants of our  method to highlight the contribution of various components of the overall technique. GRU-MP indicates that we use our two-layer GRU model, while the common `Mean Pooling (MP)' strategy is adopted to resolve the temporal dimension of videos. `C3D' and `IRV2' in the parentheses identify  the networks used to compute the visual codes. We abbreviate the joint use of C3D and IRV2 as `CI'. We use `EVE' to denote our Enriched Visual Encoding that applies Hierarchical Fourier Transform - indicated by  the subscript `hft' - on the activations of the network extraction layers. The proposed final technique, that also incorporates the high-level semantic information - indicated by the subscript `+sem' - is mentioned in the last row of the Table. We also follow the same notational conventions for our method in the remaining Tables.   

\begin{table}[t]
  \centering
  \small

 \caption{Benchmarking on MSVD dataset~\cite{chen2011collecting} in terms of BLEU-4 (B-4), METEOR (M), CIDEr$_D$ (C) and ROUGE$_L$ (R). See the text for the description of proposed method GRU-EVE's variants.}
  \begin{tabular}{|l|c|c|c|c|}
    \hline
  \textbf{Model} & \textbf{B-4} & \textbf{M} & \textbf{C} & \textbf{R} \\
    \hline
    \hline
    FGM~\cite{thomason2014integrating}                  & 13.7    &  23.9 & -     & - \\
    S2VT~\cite{venugopalan2015sequence}                 & -       & 29.2  & -     & - \\
    LSTM-YT~\cite{venugopalan2014translating}            & 33.3   & 29.1  & -     & - \\
    Temporal-Attention (TA)~\cite{yao2015describing}      & 41.9  & 29.6  & 51.67 & - \\
    h-RNN~\cite{yu2016video}                              & 49.9  & 32.6  & 65.8 & - \\
    MM-VDN~\cite{xu2015multi}                             & 37.6  & 29.0  & -     & - \\
    HRNE~\cite{pan2016hierarchical}                       & 43.8 & 33.1 & - & - \\
    GRU-RCN~\cite{ballas2015delving} & 47.9  & 31.1 & 67.8 & - \\
    LSTM-E~\cite{pan2016jointly}     & 45.3 & 31.0 & - & - \\
    %LSTM-LS (VGG19)~\cite{xxxx}      & - & - & - & - \\
    SCN-LSTM~\cite{gan2017semantic}  & 51.1 & 33.5 & 77.7 & - \\
    DMRM~\cite{yang2017catching}  & 51.1 & 33.6 & 74.8 & - \\
    LSTM-TSA~\cite{Pan_2017_CVPR}    & \textbf{52.8} & 33.5 & 74.0 & - \\
    TDDF~\cite{Zhang_2017_CVPR}      & 45.8 & 33.3 & 73.0 & 69.7 \\
    BAE~\cite{Baraldi_2017_CVPR}     & 42.5 & 32.4 & 63.5 & - \\
    PickNet~\cite{chen2018less}      & 46.1 & 33.1 & 76.0 & 69.2 \\
    aLSTMs~\cite{gao2017video}       & 50.8 & 33.3 & 74.8 & - \\
    %LSTM-CRT~\cite{xxxx}             & 46.9 & 32.6 & 70.6 & - \\
    %HBA~\cite{xxxx}                  & 42.5 & 32.4 & - & - \\
    M$^3$-IC~\cite{wang2018m3}       & \textbf{52.8} & 33.3 & - & - \\
    RecNet$_{local}$~\cite{wang2018reconstruction} & 52.3 & 34.1 & \textbf{80.3} & 69.8 \\
    \hline
    \hline
    GRU-MP - (C3D)    & 28.8 & 27.7 & 42.6 & 61.6  \\
    % GRU-MP$_{VGG16}$  & 39.4 & 30.4 & 58.3 & 65.9  \\
    GRU-MP - (IRV2)   & 41.4 & 32.3 & 68.2 & 67.6  \\
    GRU-MP - (CI)        & 41.0 & 31.3 & 61.9 & 67.6  \\
    GRU-EVE$_{\text{hft}}$ - (C3D)         & 40.6 & 31.0 & 55.7 & 67.4 \\
    GRU-EVE$_{\text{hft}}$ - (IRV2)        & 45.6 & 33.7 & 74.2 & \textbf{69.8} \\
    %G3D-FTP$_{IRV2(1&2 \\
    %layers)}$               & - & - & - & - \\
    % GRU-EVE$_{VGG16}$       & 41.2 & 31.9 & 59.7 & 68.4 \\
    % GRU-EVE$_{CI-3L}$          & 46.6 & \textbf{34.6} & 73.9 & \textbf{70.6}\\
    GRU-EVE$_{\text{hft}}$ - (CI)          & 47.8 & \textbf{34.7} & 75.8 & \textbf{71.1}\\ \hline % 2L
    %G3D-FTP$_{VI}$          & \textcolor{red}{to} & \textcolor{red}{be} & \textcolor{red}{tested} & \textcolor{red}{-} \\
    GRU-EVE$_{\text{hft+sem}}$ - (CI)         & 47.9 & \textbf{35.0} & 78.1 & \textbf{71.5} \\ % 2L
  \hline

  \end{tabular}%
  \label{tab:MSVD}%
\vspace{-4mm}
\end{table}%

% =============================================================================

Our method achieves a strong $35$ value of METEOR, which provides a $\frac{35.0 - 34.1}{34.1}\times 100 = 2.64\%$ gain over the closest competitor. Similarly,  gain over the current state-of-the-art for ROUGE$_L$ is $2.44\%$. For the other metrics, our scores  remain competitive to the best performing methods. It is emphasized, that our approach derives its main strength from the visual encoding part in contrast to sophisticated language model, which is generally the case for the existing methods. Naturally, complex language models entail difficult and computationally expensive training process, which is not a limitation of our approach.   

\begin{figure*}[t] %  figure placement: here, top, bottom, or page
  \centering
     \includegraphics[width=5.5in]{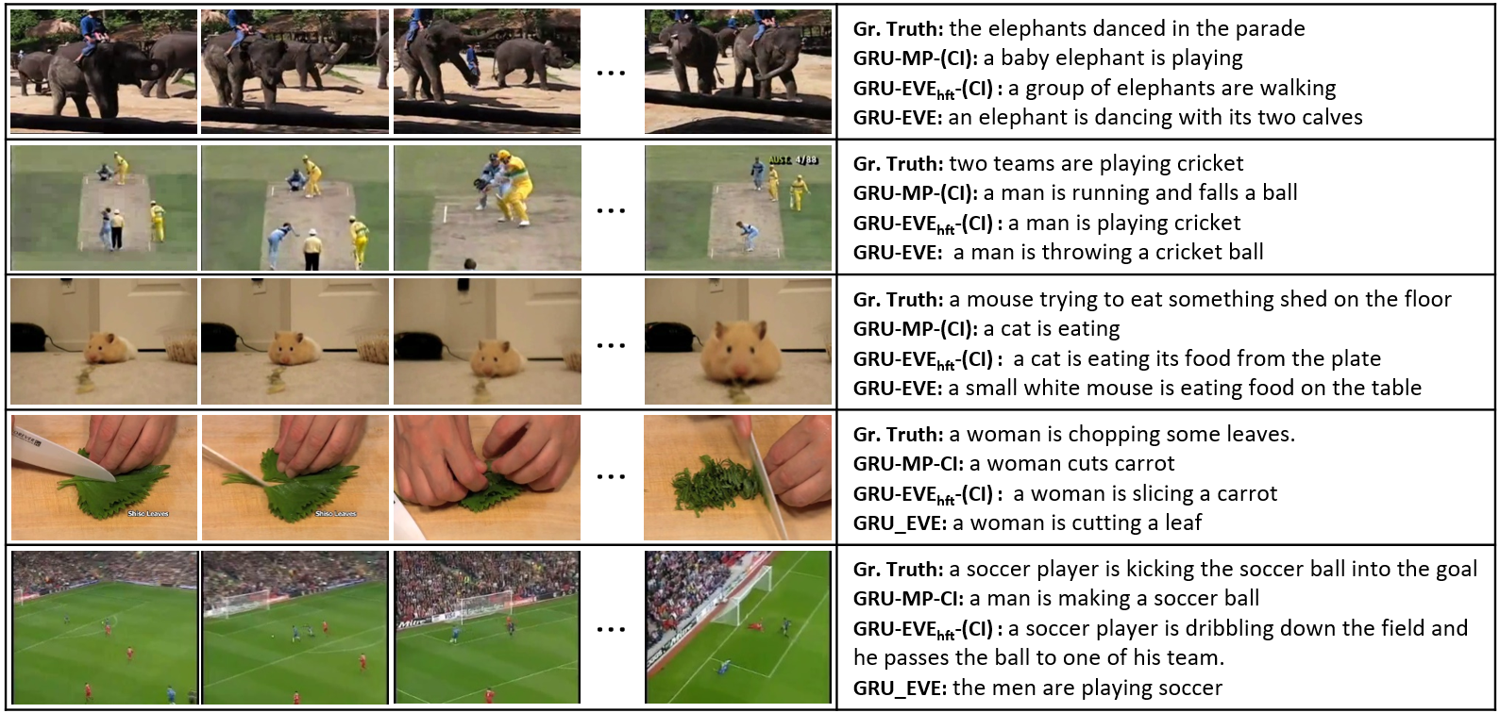}
     \caption{Illustration of caption generated for MSVD test set: The final approach is abbreviated as GRU-EVE for brevity. A sentence from ground truth captions is shown for reference.}
  \label{fig:qualanalysis}
  \vspace{-3mm}
\end{figure*}

We illustrate representative qualitative results of our method in Fig.~\ref{fig:qualanalysis}. We abbreviate our final approach as `GRU-EVE' in the figure for brevity. The semantic details and accuracy of  e.g.~plurality, nouns and verbs is clearly visible in the captions generated by the proposed method. The figure also reports the captions for GRU-MP-(CI) and GRU-EVE$_{\text{hft}}$-(CI) to show the difference resulting from hierarchical Fourier transform (hft) as compared to the Mean Pooling (MP) strategy. These captions justify the noticeable gain achieved by the proposed hft over the traditional MP in Table~\ref{tab:MSVD}. We also observe in the table that our method categorically outperforms the mean pool based methods, i.e.~LSTM-YT~\cite{venugopalan2014translating}, LSTM-E~\cite{pan2016jointly}, SCN-LSTM~\cite{gan2017semantic}, and LSTM-TSA\cite{Pan_2017_CVPR} on METEOR, CIDEr and ROUGE$_L$. Under these observations, we safely recommend the proposed hierarchical Fourier transformation as the substitute for the `mean pooling' in video captioning.    

In Table~\ref{tab:MSVD_singlefeature}, we compare the variant of our method based on a single CNN with the best performing single CNN based existing methods. The results are directly taken from~\cite{wang2018m3} for the provided METEOR metric. As can be seen, our method outperforms all these methods.  In Table~\ref{tab:MSVD_fusion}, we also compare our method on METEOR with the state-of-the-art methods that necessarily use  multiple visual features to obtain the best performance. A significant $5.1\%$ gain is achieved by our method to the closest competitor in this regard.   

%We also report that when compared with other state of the art top 5 methods (also reported from~\cite{wang2018m3}) using multiple visual features fusion on MSVD dataset, our method outperforms all of them as shown in Table~\ref{tab:MSVD_fusion}. As other METEOR score for other methods has been reported so we show only METEOR score of our method. 
%We also report the comparison of single feature based evaluation of the method in Table~\ref{tab:MSVD_singlefeature}. here we report the results based on 2D-CNN features only when passed through our neuron based temporal encoding. The scores of other methods on single feature have been reported from~\cite{wang2018m3}. Here we show that our enriched visual feature based on single 2D-CNN has outperformed all other methods as shown in Table~\ref{tab:MSVD_singlefeature}. 

\begin{table}[t]
  \centering
  \small

 \caption{Performance comparison with single 2D-CNN based methods on MSVD dataset~\cite{chen2011collecting}.}
 \begin{tabular}{|l|c|}
    \hline
  \textbf{Model} & \textbf{METEOR} \\
    \hline
    \hline
    FGM~\cite{thomason2014integrating}      & 23.90 \\
    S2VT~\cite{venugopalan2015sequence}     & 29.2 \\
    LSTM-YT~\cite{venugopalan2014translating}  & 29.07 \\
    TA~\cite{yao2015describing}             & 29.0 \\
    p-RNN~\cite{yu2016video}                & 31.1 \\
%    MM-VDN~\cite{xu2015multi}              & 37.6 & 29.0 & - & - \\
    HRNE~\cite{pan2016hierarchical}         & 33.1 \\
    BGRCN~\cite{ballas2015delving}          & 31.70 \\
    MAA~\cite{fakoor2016memory}             & 31.80 \\
    RMA~\cite{jain2017recurrent}            & 31.90 \\
    LSTM-E~\cite{pan2016jointly}            & 29.5\\
    M$^3$-inv3~\cite{wang2018m3}            & 32.18 \\
    mGRU~\cite{zhu2017bidirectional}        & 33.39 \\
    \hline
    \hline
    GRU-EVE$_{\text{hft}}$-(IRV2)           & \textbf{33.7} \\
    % GRU-EVE$_{hft}$-IRV2-2L                & - \\
    \hline
    \end{tabular}%
  \label{tab:MSVD_singlefeature}%
\end{table}%
%==============================================================================

\begin{table}[t]
  \centering
  \small

 \caption{Performance comparison on MSVD dataset~\cite{chen2011collecting}  with the methods using multiple features. The scores of existing methods are taken from \cite{wang2018m3}. V denotes VGG19, C is C3D, I$_v$ denotes  Inception-V3, G is GoogleNet and I denotes InceptionResNet-V2}
  \begin{tabular}{|l|c|}
    \hline
  \textbf{Model} & \textbf{METEOR} \\
    \hline
    \hline
    SA-G-3C~\cite{yao2015describing}             & 29.6 \\
    S2VT-RGB-Flow~\cite{venugopalan2015sequence}     & 29.8 \\
        LSTM-E-VC~\cite{pan2016jointly}            & 31.0\\
    p-RNN-VC~\cite{yu2016video}                & 32.6 \\
    M$^3$-I$_v$C~\cite{wang2018m3}                  & 33.3\\
    \hline
    \hline
    % GRU-EVE$_{hft}$-CIr-3L        & \textbf{34.6} \\
    %GRU-EVE$_{hft}$ - (CI)        & \textbf{34.7} \\ %2
    GRU-EVE$_{\text{hft+sem}}$ - (CI)        & \textbf{35.0} \\ %2
    \hline
    \end{tabular}%
  \label{tab:MSVD_fusion}%
\vspace{-5mm}
\end{table}%
% =========================================================================
% =========================================================================

%As shown in Table~\ref{tab:MSVD}, when we encode each neuron features into time domain, there is significant improvement in the results of all metrics. For instance, mean pool score for METEOR in our framework is 31.3 and when we perform the neural temporal encoding kepping the same architecture, the score improves to 34.7 making $\dfrac{34.7-31.3}{31.3} = 10.86\%$ improvement. Similarly CIDEr score improves by $\dfrac{75.8-61.9}{61.9} = 22.45\%$. Similarly there is improvement in BLEU@4 and ROUGE$_L$ scores also by 16.58\% and 4.9\% respectively. 

%From these results it can be concluded that better performance of the proposed model benefits from temporally encoding each neuron features instead of using high level features itself or mean pooling. From the Table~\ref{tab:MSVD} we can see that we outperform all methods in METEOR and ROUGE$_L$ scores irrespective they use mean pool features or not. 
%Further, methods built on top of mean pooled features of videos e.g. LSTM-YT~\cite{venugopalan2014translating}, LSTM-E~\cite{pan2016jointly}, SCN-LSTM~\cite{gan2017semantic}, LSTM-TSA\cite{Pan_2017_CVPR}, our method outperform all of them in three metrics METEOR, CIDEr and ROUGE$_L$. One of the reason of low BLEU-4 score in our framework is use of multi layer GRU that we discuss later.

\vspace{-2mm}
\subsubsection{Results on MSR-VTT dataset}
% \vspace{-1mm}
MSR-VTT~\cite{xu2016msr} is a recently released dataset. We compare performance of our approach on this dataset with the latest published models such as Alto~\cite{shetty2016frame}, RUC-UVA~\cite{dong2016early}, TDDF~\cite{Zhang_2017_CVPR}, PickNet~\cite{chen2018less}, M$^3$-VC~\cite{wang2018m3} and RecNet$_{local}$~\cite{wang2018reconstruction}. The results are summarized in Table~\ref{tab:MSR-VTT}. 
Similar to the MSVD dataset, our method significantly improves the state-of-the-art on this dataset on METEOR and ROUGE$_L$ metrics, while achieving strong results on the remaining metrics. 
These result ascertain the effectiveness of the proposed enriched visual encoding for visual captioning. We provide examples of qualitative results on this dataset in the supplementary material of the paper.

\begin{table}[t]
  \centering
  \small
 \caption{Benchmarking on MSR-VTT dataset~\cite{xu2016msr} in terms of BLEU-4 (B-4), METEOR (M), CIDEr$_D$ (C) and ROUGE$_L$ (R).}
  \begin{tabular}{|l|c|c|c|c|}
    \hline
  \textbf{Model} & \textbf{B-4} & \textbf{M} & \textbf{C} & \textbf{R} \\
    \hline
    Alto~\cite{shetty2016frame}         & 39.8 & 26.9 & 45.7 & 59.8 \\
    RUC-UVA~\cite{dong2016early}        & 38.7 & 26.9 & 45.9 & 58.7 \\
    TDDF~\cite{Zhang_2017_CVPR}         & 37.3  & 27.8 & 43.8 & 59.2 \\
    PickNet~\cite{chen2018less}         & 38.9 & 27.2 & 42.1 & 59.5 \\
    M$^3$-VC~\cite{wang2018m3}          & 38.1 & 26.6 & - & - \\
    RecNet$_{local}$~\cite{wang2018reconstruction} & 39.1 & 26.6 &  42.7 & 59.3 \\
    \hline
    \hline
    GRU-EVE$_{\text{hft}}$ - (IRV2)       & 32.9 & 26.4 & 39.2 & 57.2 \\ % (3 Layer)
    % GRU-EVE$_{\text{hft}}$ - (CI)         & 35.7 & 27.1 & 43.1 & 59.1  \\ % (3 Layer)
    GRU-EVE$_{\text{hft}}$ - (CI)         & 36.1 & 27.7 & 45.2 & 59.9 \\ % (2 Layer)
    GRU-EVE$_{\text{hft+sem}}$ - (CI)     & 38.3 & \textbf{28.4} & \textbf{48.1} & \textbf{60.7} \\ 
    \hline
    \end{tabular}%
  \label{tab:MSR-VTT}%
\vspace{-4mm}
\end{table}%

% ==============================================================

\vspace{-1mm}
\section{Discussion}
\label{sec:Disc}
\vspace{-1mm}
We conducted a through empirical evaluation of the proposed method to explore its different aspects. Below we discuss and highlight few of these aspects in the text. Where necessary, we also provide results in the supplementary material of the paper to back the discussion.    

For the settings discussed in the previous section, we generally observed semantically rich captions generated by the proposed approach. In particular, these captions well captured the plurality of objects and their motions/actions. Moreover, the captions generally described the whole videos instead of its partial clips. Instead of only two, we also tested different number of GRU layers, and observed that  increasing the number of GRU layers deteriorated the BLEU-4 score. However, there were improvements in all the remaining metrics. We retained only two GRU layers in the final method mainly for computational gains. Moreover, we also tested different architectures of GRU, e.g.~with state sizes 512, 1024, 2048 and 4096. We observed a trend of performance improvement until 2048 states. However, further states did not improve the performance. Hence, 2048 were finally used in the results reported in the previous section. 

Whereas all the components of the proposed technique contributed to  the overall final performance, the biggest revelation of our work is the use of hierarchical Fourier Transform to capture the temporal dynamics of videos. As compared to the `nearly standard' mean pooling operation performed in the existing captioning pipeline, the proposed use of Fourier Transform promises a  significant performance gain for any method. Hence, we safely recommend replacing the mean pooling operation with our transformation for the future techniques.

\vspace{-3mm}
\section{Conclusion}
\vspace{-2mm}
We presented a novel technique for visual encoding of videos to generate semantically rich captions. Besides capitalizing on the representation power of CNNs, our method explicitly accounts for the spatio-temporal dynamics of the scene, and high-level semantic concepts encountered in the video. We applying Short Fourier Transform to 2D and 3D CNN features of the videos in a hierarchical manner, and account for the high-level semantics by processing output layer features of an Object Detector and the 3D CNN. Our enriched visual representation is used to learn a relatively simple GRU based language model that performs on-par or better than   the existing video description methods on popular MSVD and MSR-VTT datasets. 

\noindent \textbf{Acknowledgment} This research was supported by ARC Discovery Grant DP160101458 and partially by DP190102443. The Titan XP GPU used in our experiments was donated by NVIDIA corporation.

{\small
\bibliographystyle{ieee}
\bibliography{Draft_paper_final}
}

\end{document}